\documentclass[sigconf, nonacm]{acmart}

\AtBeginDocument{%
  }

\copyrightyear{2025}
\acmYear{2025}
\setcopyright{none}
\acmConference[arXiv preprint]{arXiv preprint}{2025}{}
\acmBooktitle{}
\acmDOI{}
\acmISBN{}
\renewcommand\footnotetextcopyrightpermission[1]{} 

\usepackage{multirow}
\usepackage{enumitem}
\usepackage[skip=2pt]{caption}
\usepackage{bm}
\begin{document}

\title{FoodLogAthl-218: Constructing a Real-World Food Image Dataset\\ Using Dietary Management Applications}

\author{Mitsuki Watanabe}
\orcid{0009-0009-7616-3053}
\thanks{This is the authors' preprint version of the paper accepted to ACM Multimedia 2025 (Dataset Track).
 The official version is available at: \url{https://doi.org/10.1145/3746027.3758276}}
\affiliation{%
  \institution{The University of Tokyo}
  \city{Tokyo}
  \country{Japan}}
\email{watanabe@hal.t.u-tokyo.ac.jp}

\author{Sosuke Amano}
\orcid{0000-0001-7463-2631}
\affiliation{%
  \institution{foo.log Inc.}
  \city{Tokyo}
  \country{Japan}}
\email{amano@foo-log.co.jp}

\author{Kiyoharu Aizawa}
\orcid{0000-0003-2146-6275}
\affiliation{%
  \institution{The University of Tokyo}
  \city{Tokyo}
  \country{Japan}}
\email{aizawa@hal.t.u-tokyo.ac.jp}

\author{Yoko Yamakata}
\orcid{0000-0003-2752-6179}
\affiliation{%
  \institution{The University of Tokyo}
  \city{Tokyo}
  \country{Japan}}
\email{yamakata@hal.t.u-tokyo.ac.jp}

\renewcommand{\shortauthors}{Mitsuki Watanabe, Sosuke Amano, Kiyoharu Aizawa, and Yoko Yamakata}

\renewcommand\UrlFont{\ttfamily\small}

\begin{abstract}

Food image classification models are crucial for dietary management applications because they reduce the burden of manual meal logging. However, most publicly available datasets for training such models rely on web-crawled images, which often differ from users' real-world meal photos. In this work, we present \textit{FoodLogAthl-218}, a food image dataset constructed from real-world meal records collected through the dietary management application FoodLog Athl.
The dataset contains 6,925 images across 218 food categories, with a total of 14,349 bounding boxes.
Rich metadata, including meal date and time, anonymized user IDs, and meal-level context, accompany each image.
Unlike conventional datasets—where a predefined class set guides web-based image collection—our data begins with user-submitted photos, and labels are applied afterward. This yields greater intra-class diversity, a natural frequency distribution of meal types, and casual, unfiltered images intended for personal use rather than public sharing.
In addition to (1) a standard classification benchmark, we introduce two FoodLog-specific tasks: 
(2) an incremental fine-tuning protocol that follows the temporal stream of users' logs, and 
(3) a context-aware classification task where each image contains multiple dishes, and the model must classify each dish by leveraging the overall meal context.
We evaluate these tasks using large multimodal models (LMMs). The dataset is publicly available at \url{https://huggingface.co/datasets/FoodLog/FoodLogAthl-218}.

\end{abstract}



\begin{teaserfigure}
  \centering
  \includegraphics[width=\linewidth]{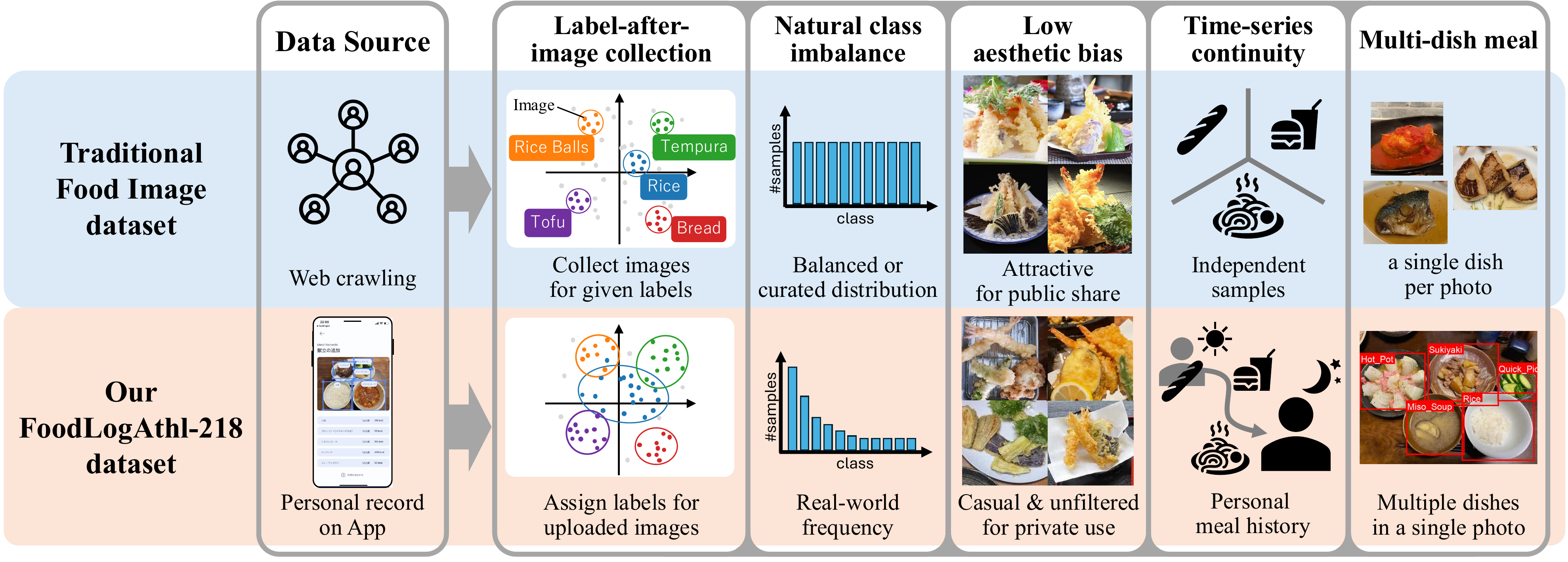}
  \caption{Comparison between traditional food image datasets and FoodLogAthl-218.}
  \label{distribution}
\end{teaserfigure}

\maketitle

\section{Introduction}

\begin{table*}[thb]
\centering
\caption{Comparison of Datasets for Food Image Classification Tasks.}
\begin{tabular}{@{}lrrrrr@{}}
\toprule
Year & Dataset Name & \#classes & \#samples & Cuisine Category & Data Source \\
\midrule
2009 & PFID~\cite{chen2009pfid} & 101 & 4,545 & Western & Fast Food Chains \\
2014 & UEC-Food 256~\cite{kawano2015automatic} & 256 & 25,088 & International & \textbf{Web} + Original Database \\
2014 & Food-101~\cite{bossard2014food} & 101 & 101,000 & Western & \textbf{Web} (foodspotting.com) \\
2015 & UPMC Food-101~\cite{wang2015recipe} & 101 & 90,840 & Western & \textbf{Web} (Google~\cite{google}) \\
2016 & UNICT-FD1200~\cite{farinella2016retrieval} & 1,200 & 4,754 & International & Experiment Participants \\
2017 & Vegfru~\cite{hou2017vegfru} & 25/292\textsuperscript{*} & 160,000 & Vegetables, Fruits & \textbf{Web} (Google, ImageNet~\cite{imagenet}, etc.) \\
2019 & FoodAI-756~\cite{sahoo2019foodai} & 152/756\textsuperscript{*} & 400,000 & International & \textbf{Web} (Google, Bing, etc.) \\
2021 & Food2K~\cite{min2023large} & 2,000 & 1,036,564 & International & \textbf{Web} (Meituan~\cite{meituan}) \\
2023 & Food-500 Cap~\cite{ma2023food} & 494 & 24,700 & International & \textbf{Web} (Google, Bing) \\
\midrule
2025 & FoodLogAthl-218 & 218 & 14,349 & International & \textbf{Dietary Management App} \\
\bottomrule
\end{tabular}
\label{Food image dataset table}

\begin{itemize}[leftmargin=7.2em]
\setlength{\parskip}{0cm} 
\setlength{\itemsep}{0cm} 
\item[*] Hierarchically structured dataset, where $n$/$m$ indicates $m$ classes divided into $n$ higher-level classes.
\end{itemize}
\end{table*}

Most publicly available food-category classification datasets are built from web-sourced images~\cite{joutou2009food,hoashi2010image,matsuda2012multiple,anthimopoulos2014food,kawano2015automatic,bossard2014food,wang2015recipe,zhou2016fine,hou2017vegfru,qiu2022mining,kaur2019foodx,min2019ingredient,sahoo2019foodai,min2020isia,min2023large,bender2023learning,ma2023food}.
Although some studies utilize proprietary food logs, these datasets are typically not publicly available due to privacy and licensing constraints~\cite{Aizawa2013FoodBalance,Horiguchi2018Personalized,Kim2021Boosting}.

To address this gap, we constructed and publicly released a new food image dataset, \textbf{FoodLogAthl-218}, based on users’ personal food records collected via a food management application we developed. The dataset was curated with approval from our institutional ethics board; informed consent was obtained from all participants, and personally identifiable information was removed via a two-stage manual process.
Since the raw user-submitted data included missing or incorrect detections and noisy labels, we applied a custom multi-stage filtering pipeline to produce a clean, research-ready subset consisting of 6,925 images and 14,349 bounding boxes across 218 food categories—comparable in scope to existing classification datasets.

The \textbf{FoodLogAthl-218} dataset introduces the following five key properties that distinguish it from existing food image benchmarks and make it better suited for real-world dietary logging scenarios.

\noindent{\bf Label-after-image collection} 
Unlike most existing datasets, where a fixed set of class labels is defined first, and images are then collected to match those labels, our dataset begins with user‐submitted meal photos, which are labeled post hoc. This bottom-up process yields significantly higher intra-class diversity and more accurately reflects the nature of real-world food logging tasks.

\noindent{\bf Natural class imbalance} While existing datasets often attempt to balance the number of samples per class or rely on the availability of web-sourced images, our dataset preserves the natural frequency distribution of meals as they appear in everyday food records.

\noindent{\bf Low aesthetic bias} 
Unlike web‐sourced images for public presentation (e.g., recipes, advertisements, or social media), our dataset consists of casual, unfiltered photos taken for personal use. These images often feature poor lighting, unstyled plating, or partially eaten dishes—scenarios rarely encountered in web‐collected data, and thus exhibit a fundamentally different visual character.

\noindent{\bf Time-series continuity for personalization} 
Since many users log meals over extended periods, the dataset has a partial time‐series structure. This temporal continuity facilitates
longitudinal modeling and enables evaluation of incremental learning and personalization.

\noindent{\bf Multi-dish meal composition} 
Each image in our dataset may contain multiple dishes that co‐occur on the same plate or tray, reflecting realistic meal compositions and enabling context‐aware classification based on dish co‐occurrence within a single meal.

Figure~\ref{distribution} illustrates several of these distinguishing properties, including visual diversity, meal composition, and real-world data characteristics. Taken together, these aspects highlight the limitations of conventional food image datasets in supporting practical food-logging applications and demonstrate the need for more realistic, user-centered resources like FoodLogAthl-218.

\section{Related Works}

\subsection{Image-Based Food Logging}
\label{RelatedWorks_FoodLoggingByFoodImage}

With the widespread adoption of smartphones, many meal management applications have been released. In Japan, Asken~\cite{asken}, MyFitnessPal~\cite{myfitnesspal}, and Calomeal~\cite{karo} are prominent. These apps utilize image recognition technology to automatically recognize the content of a meal and estimate its nutritional value. 

In this research, we used the food recording app, ``FoodLog Athl''~\cite{nakamoto2022foodlog} developed by our laboratory for academic purposes. One key feature of FoodLog Athl is its ability to facilitate communication between users and registered dietitians. 
Once users link their account on the app, they can share their food records with an assigned dietitian, who can then review the records and provide personalized nutritional guidance.
To ensure accurate nutritional analysis, FoodLog Athl does not allow users to enter free‐text dish names. Instead, users select dishes from a predefined database that contains approximately 1,400 common Japanese meals and an additional 140K items (including everyday foods, restaurant dishes, and packaged products). This database is compiled from SARAH~\cite{SARAH} and the Standard Tables of Food Composition in Japan~\cite{STFCJ}.

\begin{table*}[tb]
  \centering
  \caption{Data Filtering Process and the Number of Images, Bounding Boxes (bboxes) and Classes.}
  \begin{tabular}{@{} l l
                  r @{ (} r @{)}
                  r @{ (} r @{)}   
                  r @{ (} r @{)}   
                @{}}
    \toprule
    Step & Operation & \multicolumn{1}{r@{\phantom{ (}}}{\#images} & \multicolumn{1}{r@{}}{(diff)} & \multicolumn{1}{r@{\phantom{ (}}}{\#bboxes} & \multicolumn{1}{r@{}}{(diff)} & \multicolumn{1}{r@{\phantom{ (}}}{\#classes} & \multicolumn{1}{r@{}}{(diff)}  \\
    \midrule
    0 & Raw data
      & \multicolumn{1}{r@{\phantom{ (}}}{11,460}  
      & \multicolumn{1}{r@{}}{\phantom{)}}
      & \multicolumn{1}{r@{\phantom{ (}}}{61,414}  
      & \multicolumn{1}{r@{}}{\phantom{)}}        
      & \multicolumn{1}{r@{\phantom{ (}}}{5,889}  
      & \multicolumn{1}{r@{}}{\phantom{)}}        \\
    1 & Remove samples w/o image              &11,460&0& 26,646 & -34,768 & 3,167  & -2,722 \\
    2 & Remove samples w/o bounding box       &11,285&-175& 26,065 & -581    & 3,071  & -96    \\
    3 & Remove classes w/ <10 samples         &9,121&-2,164& 21,944 & -4,121  & 1,180  & -1,891 \\
    4 & Remove outliers based on image features &7,537&-1,584& 17,513 & -4,431  & 1,180  & 0      \\
    5 & Manual review and remove invalid images &6,925&-612& 14,349 & -3,164    & 1,180    & 0     \\
    6 & Merge similar classes and clean labels &6,925&0& 14,349 & 0       & 218    & -982   \\
    \bottomrule
  \end{tabular}
  \label{data cleaning steps table}
\end{table*}

\subsection{Food Image Datasets}
\label{FoodImageDatasets}

For models to accurately 
predict real-world data, they must be trained on rich data that reflects real-world environments. Thus, datasets serve as the foundation for training and evaluating models, with their quality, scale, and diversity directly linked to model performance. In the field of dietary studies, research on food image datasets has progressed, with datasets such as Food-101~\cite{bossard2014food} containing 101,000 images across 101 classes, UEC-Food 256~\cite{kawano2015automatic} containing 25,088 images across 256 classes, and Food2K~\cite{min2023large} containing 1,036,564 images across 2,000 classes, reflecting the growing scale of these datasets over the years.

As seen in Table~\ref{Food image dataset table}, most existing food image datasets constructed for classification tasks rely on the Web, including social networking sites (SNS), as their primary data source. 

\subsection{Vision-Language Models in the Food Domain}
\subsubsection{CLIP}

Contrastive Language–Image Pretraining (CLIP)~\cite{CLIP} serves as a baseline for food image classification thanks to its strong zero-shot accuracy. For example, CLIP’s zero-shot classifier achieves over 90\% Top-1 accuracy on Food-101~\cite{CLIP}—outperforming a ResNet-50–based classifier~\cite{He2016-zx} by more than 15 points. Moreover, in ingredient estimation tasks, CLIP-based models yield higher per-ingredient recall and F1 scores than ResNet-based counterparts~\cite{10.1145/3664647.3684997}.

\subsubsection{Large Multi‐modal Models (LMMs)}

Beyond CLIP, recent progress in multimodal learning has centered on general‐purpose LMMs, which combine image encoders and large language models into end‐to‐end frameworks. Models such as LLaVA~\cite{liu2024visual}, MiniGPT-4~\cite{zhu2023minigpt}, OpenFlamingo~\cite{awadalla2023openflamingo}, Phi-3.5-Vision~\cite{abdin2024phi}, and Qwen-VL~\cite{bai2023qwen} have all demonstrated strong zero‐shot and few-shot performance across a range of food‐related tasks. For example, FoodMLLM-JP leverages these architectures for Japanese recipe generation with excellent fluency and culinary accuracy~\cite{Imajuku2025FoodMLLM}.  
Domain‐adapted variants—such as FoodLMM~\cite{yin2023foodlmm}, ChefFusion~\cite{li2024cheffusion}, and RoDE~\cite{jiao2024rode}—further boost fine‐grained classification, ingredient localization, and nutrition estimation.  
These successes confirm that CLIP and LMMs serve as robust baselines for our food image benchmarks.

\begin{figure}[tb]
  \centering
  \includegraphics[width=\linewidth]{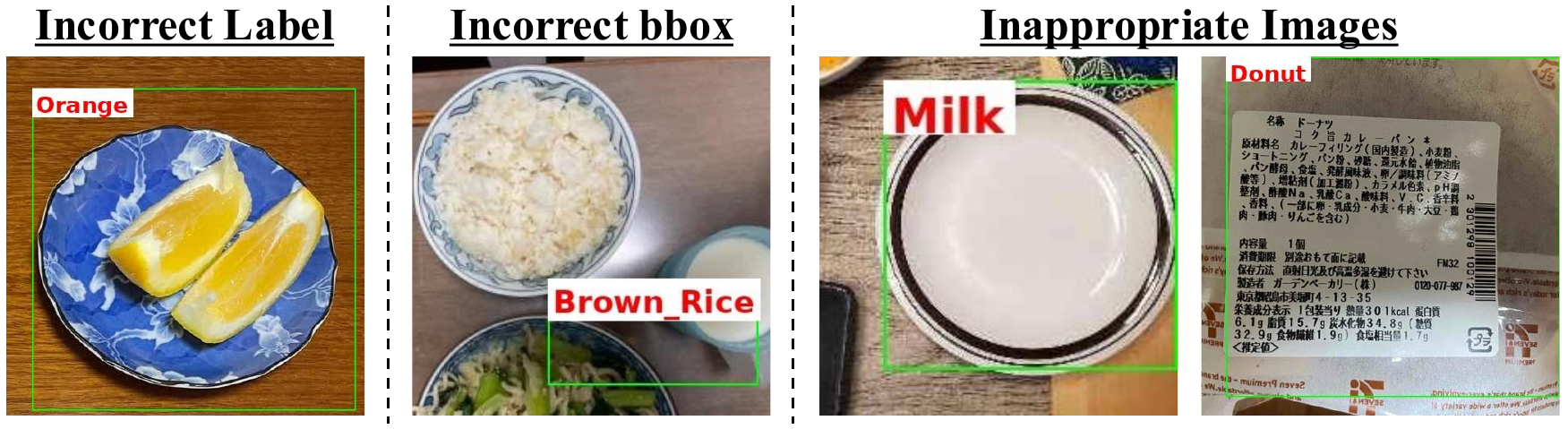}
  \caption{Examples of erroneous data. }
  \label{erroneous_data}
\end{figure}

\section{FoodLogAthl-218 Dataset}

As mentioned in the Introduction, food records submitted by users in daily food-logging applications tend to be highly noisy, as shown in Figure~\ref{erroneous_data}. To address this, we developed a custom multi-stage filtering pipeline to extract a clean subset comparable in quality to existing food image datasets.
\subsection{Data Collection}
\label{DataCollection}

The dataset used in this study consists of food records collected via the FoodLog Athl app from May 2023 to October 2024, covering 632 users. 
Since many users upload only once, or submit non-food images, we applied the following filtering steps to obtain a reliable subset of approximately 60,000 records: 
(i) selected dietitians linked to at least three user accounts, 
(ii) identified all user accounts associated with those dietitians, and 
(iii) retrieved every meal record from those users. 
Each record contains an anonymized user ID, meal date and time, dish name, portion size, an image with bounding box annotations, and nutritional values (all initially inferred by the app’s built-in detection and recognition models and then corrected by users as needed). 
As images often contain multiple dishes, each dish is treated as an individual sample, allowing a single image to correspond to multiple dataset samples.

\begin{table*}[tb]
  \centering
  \caption{Comparison of distribution metrics between FoodLogAthl-218 and UEC-Food 256 using feature extractors. Higher \emph{Trace} and \emph{Davies--Bouldin} indicate greater spread, while lower \emph{Silhouette} and \emph{Calinski--Harabasz} indicate less compact clusters.}
  \resizebox{\textwidth}{!}{%
    \begin{tabular}{@{}lcccccccccccc@{}}
      \toprule
        & \multicolumn{3}{c}{\textbf{Trace (\(\uparrow\))}}
        & \multicolumn{3}{c}{\textbf{Silhouette (\(\downarrow\))}}
        & \multicolumn{3}{c}{\textbf{Davies--Bouldin (\(\uparrow\))}}
        & \multicolumn{3}{c}{\textbf{Calinski--Harabasz (\(\downarrow\))}} \\
      \cmidrule(lr){2-4} \cmidrule(lr){5-7} \cmidrule(lr){8-10} \cmidrule{11-13}
      \textbf{Dataset}
      & \textbf{CLIP} & \textbf{EVA-CLIP} & \textbf{Swin}
      & \textbf{CLIP} & \textbf{EVA-CLIP} & \textbf{Swin}
      & \textbf{CLIP} & \textbf{EVA-CLIP} & \textbf{Swin}
      & \textbf{CLIP} & \textbf{EVA-CLIP} & \textbf{Swin} \\
      \midrule
      UEC-Food 256
        & 71.21            & 45.09            & 264.00
        & 0.0335           & 0.0577           & -0.0064
        & 3.64             & 3.02             & 4.11
        & 74.05            & 115.80           & 49.17            \\
        FoodLogAthl-218 (Ours)
        & \textbf{81.70}   & \textbf{49.92}   & \textbf{282.57}
        & \textbf{-0.0210} & \textbf{0.0036}  & \textbf{-0.0288}
        & \textbf{4.52}    & \textbf{3.68}    & \textbf{4.51}
        & \textbf{27.51}   & \textbf{51.51}   & \textbf{21.29}   \\
      \bottomrule
    \end{tabular}%
  }
  \label{table:diversity_metrics_comparison}
\end{table*}

\subsection{Data Filtering}

An overview of our data cleaning steps and their impact on sample and class counts is shown in Table~\ref{data cleaning steps table}.
Steps 1 and 2 remove records with missing or incorrectly assigned images and bounding boxes. 
Step 3 drops underrepresented classes that are unsuitable for training.

Step 4 applies CLIP-based outlier detection to catch subtler errors such as mislabeled crops or non-food images.
Each dish crop is encoded with a pre-trained CLIP model (ViT-B/16), and its feature distance from the class centroid is calculated. To choose an appropriate cutoff, we manually inspected samples across different distance ranges and found that flagging those more than 1.5 standard deviations above the class mean removed clear errors with high precision while preserving valid but uncommon cases. Accordingly, any sample whose distance exceeds this threshold is flagged as an outlier.

Step 5 manually reviews all remaining samples, correcting bounding boxes and labels or removing any invalid entries to ensure final dataset quality.

To preserve the natural distribution of food records in real-world logging, we treat each image as an atomic unit.

Since one image may contain multiple dishes, removing individual samples could distort meal-level patterns. 

Instead, if any sample from an image is flagged, all associated samples are discarded. 

This all-or-none policy maintains the original frequency of multi-dish meals and supports full-meal tasks, such as the description-guided classification in Section~\ref{4-3}.
A total of 6,925 meal images with 14,349 bounding boxes from 513 users remained after filtering up to Step 5.

\subsection{Clustering}

As explained in Sec.~\ref{RelatedWorks_FoodLoggingByFoodImage}, food names (i.e., class labels) are drawn from over 140K choices and still exhibit various notation inconsistencies after filtering. 
To compare our dataset with existing food image datasets, we adopted a two-stage clustering procedure in Step 6 to merge labels referring to the same dish.
First, we prompted GPT-4o~\cite{achiam2023gpt} with ``Please group the following dish names into appropriate categories,'' leveraging its pretrained language and domain knowledge to automatically cluster synonyms, spelling variants, and minor phrasing differences. This model-assisted step greatly reduces the manual effort needed to unify a large, noisy label set. Second, we manually reviewed and refined these clusters to correct any misgroupings and ensure that each class retained clear, distinct definitions. By combining scalable, automated grouping with targeted human curation, we achieve both efficiency and high labeling accuracy in our final class definitions.

\subsection{Dataset Overview and Class Distribution}

\begin{figure}[tb]
  \centering
  \includegraphics[width=\linewidth]{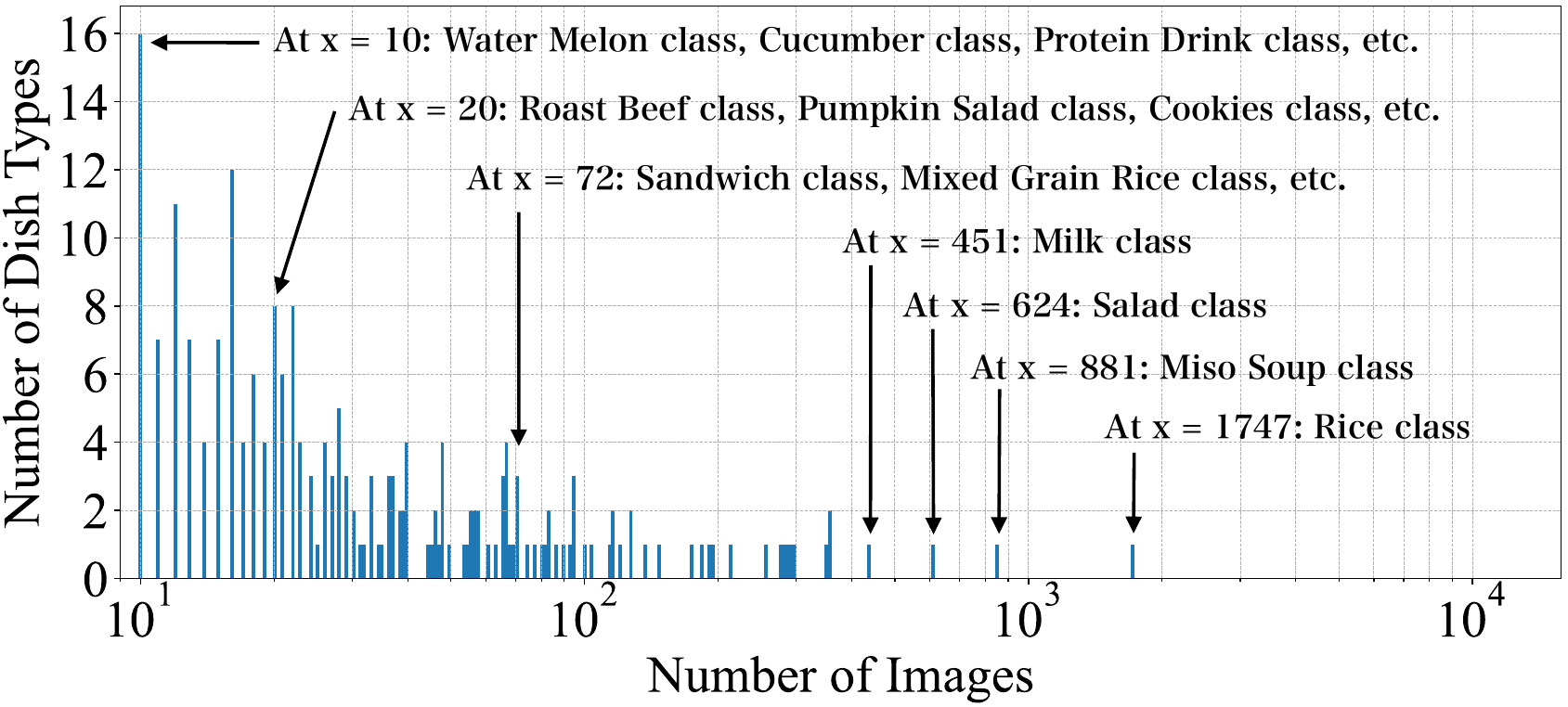}
  \caption{Log–linear plot of the number of entries per class in FoodLogAthl-218 (logarithmic horizontal axis, linear vertical axis), with illustrative examples.}
  \label{histogram}
\end{figure}

Figure~\ref{histogram} plots the distribution of entries per class in FoodLogAthl-218 on a log–linear scale (logarithmic horizontal axis, linear vertical axis). The number of samples per class is highly imbalanced, faithfully reflecting the true frequency distribution of daily meals logged by Japanese FoodLog Athl users.

\begin{figure}[tb]
  \centering
  \includegraphics[width=\linewidth]{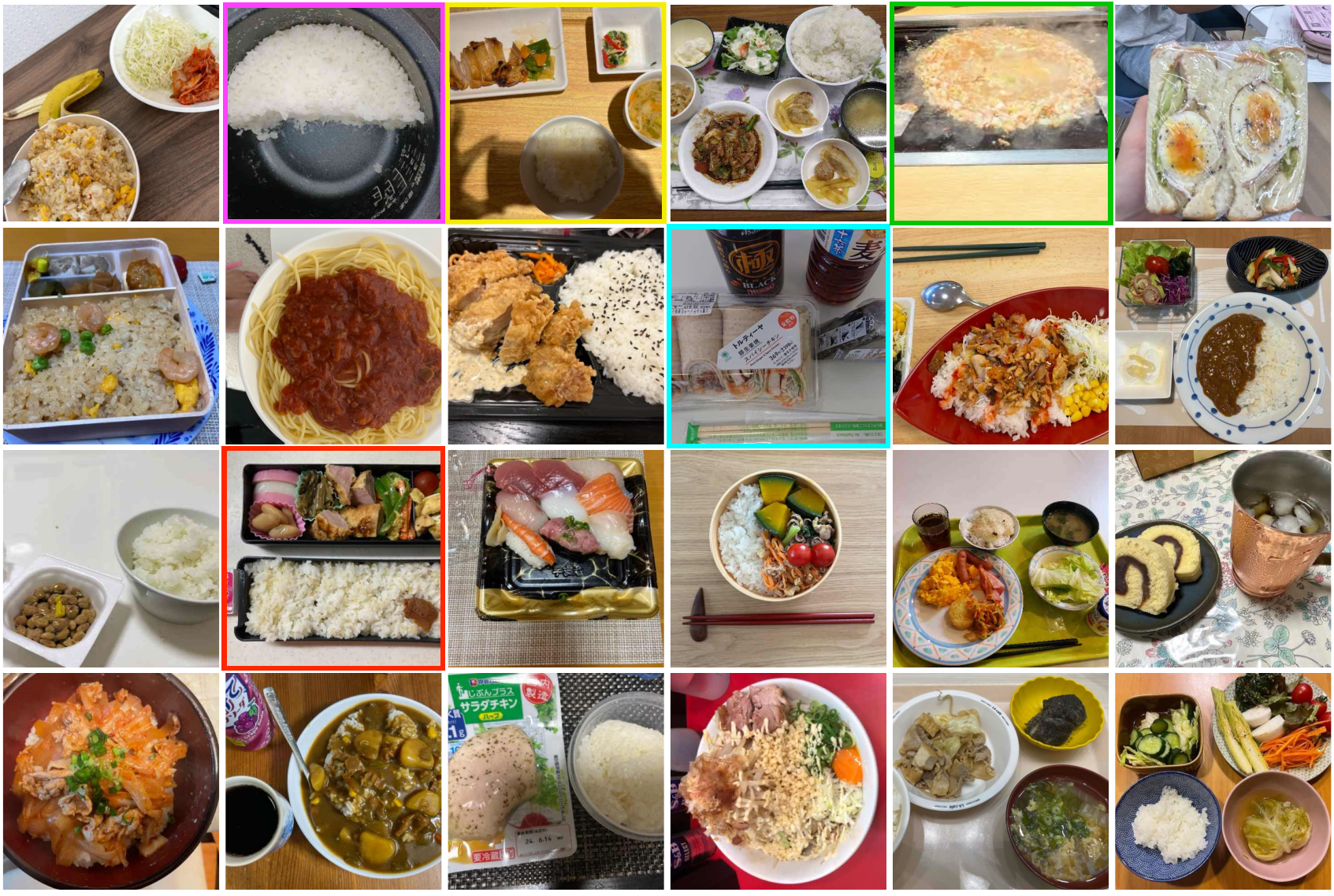}
  \caption{Examples of data included in FoodLogAthl-218.}
  \label{example}
\end{figure}

Figure~\ref{example} shows example meal images from FoodLogAthl-218. 
The dataset includes images that are rarely found on the web but commonly appear in personal food logs, such as boxed meals (red box) and packaged products purchased at convenience stores (blue box), as well as partially eaten rice in a rice cooker (purple box).
Another notable feature is the wide variation in composition and lighting. For example, in the image outlined in yellow, the photographer’s smartphone casts a shadow over the food—a common characteristic of casual, unedited meal snapshots.

To evaluate within‐class variability and between‐class separability, we extracted image features from every sample in FoodLogAthl‐218 and UEC-Food 256-another Japanese food image classification dataset- using three different encoders:
CLIP (ViT-L/14)~\cite{CLIP}, 
EVA-CLIP (\texttt{eva\_giant\_patch14\_clip\_224})~\cite{sun2023eva}, and 
Swin Transformer (\texttt{swin\_large\_patch4\_window12\_384.ms\_in22k})~\cite{liu2021swin}.
For each encoder and its corresponding feature set, we computed four dispersion metrics to evaluate class separability and variability:

\begin{itemize}[leftmargin=*]
  \item \textbf{Trace} ($\uparrow$)~\cite{jolliffe2002pca}: Sum of variances across all feature dimensions.  Higher values indicate more within‐class spread.
  \item \textbf{Silhouette Score} ($\downarrow$)~\cite{rousseeuw1987silhouettes}: Measures class cohesion and separation; lower scores suggest greater overlap.
  \item \textbf{Davies–Bouldin Index} ($\uparrow$)~\cite{davies1979cluster}: Ratio of within‐class scatter to between‐class separation.  Larger values reflect less compact, more overlapping classes.
  \item \textbf{Calinski–Harabasz Score} ($\downarrow$)~\cite{calinski1974dendrite}: Ratio of between‐class to within‐class variance. Lower values indicate weaker separation.
\end{itemize}

Table~\ref{table:diversity_metrics_comparison} summarizes these four metrics for each encoder.  The arrows in the column headers denote the direction ($\uparrow$ larger, $\downarrow$ smaller) that corresponds to higher intra‐class variance and inter‐class overlap, i.e., greater difficulty of classification. Across all metrics and encoders, FoodLogAthl‐218 shows consistently higher intra‐class variance and inter‐class overlap than UEC-Food 256, demonstrating that our user‐submitted, bottom‐up collection yields more visual diversity (and noise) than datasets gathered via traditional web‐based methods.

\section{Benchmark Tasks and Experimental Setup }
\begin{figure}[tb]
  \centering
  \includegraphics[width=\linewidth]{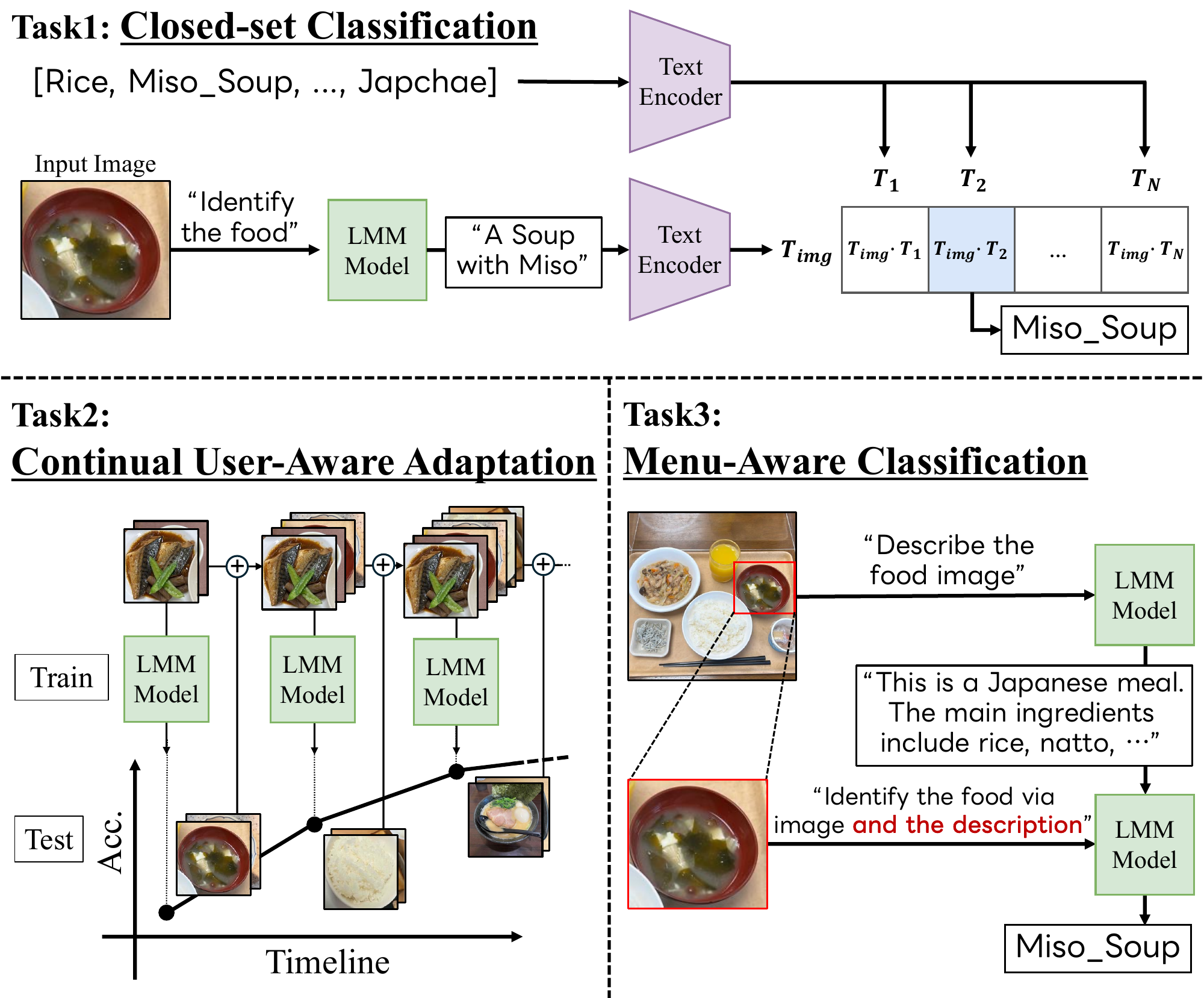}
  \caption{Overview of the three evaluation tasks.}
  \label{task}
\end{figure}

In this section, we define three tasks to evaluate the complexity of the FoodLogAthl-218 dataset and provide baseline methods for each using large multimodal models (LMMs), as illustrated in Figure~\ref{task}.

\subsection{Task 1. Standard Closed-Set Classification}  
\label{4-1}
\subsubsection{Task Definition and baseline method}
As in conventional food-image benchmarks, each dish crop is assigned exactly one of the 218 class labels. 
We benchmarked this task across various LMMs using \texttt{lmms-eval}, a unified and standardized multimodal evaluation framework that includes more than 10 models\cite{zhang2024lmmsevalrealitycheckevaluation,lmms_eval2024}.
In the baseline method, we first generated a free-form description for each dish image crop, then embedded both the description and all 218 class labels using the \texttt{intfloat/e5-large-v2} text encoder~\cite{wang2023e5}. The predicted label was selected as the one with the highest cosine similarity to the description embedding. This overall pipeline is illustrated at the top of Figure~\ref{task}.
For EVA-CLIP and CLIP, classification was performed by directly computing cosine similarity between the image features and label text embeddings.
We report Top-1/3/5 accuracy to account for uncertainty and label ambiguity.

\subsubsection{Experimental Results}

Table~\ref{table:classification_comparison} shows that GPT-4o achieves the highest recall@1 ($\approx$42\%), and among all open-source models Qwen2.5-VL-7B-Instruct is strongest, but even the best LMMs stall around 40\% recall@1, highlighting the challenge of FoodLogAthl-218.
Interestingly, a smaller-scale classification model compared to LMMs, CLIP (ViT-L/14) attains the top recall@5 (60\%). This stems from its direct image–text similarity scoring, which naturally produces a full ranked list, whereas embedding-based LMM descriptions tend to concentrate probability on their first guess and are less effective at generating diverse Top-N candidates.

\begin{table}[tb]
  \centering
  \caption{Classification performance on FoodLogAthl-218.}
  \resizebox{\columnwidth}{!}{%
    \begin{tabular}{@{}lcccc@{}}
      \toprule
      \textbf{Model} & \textbf{F1} & \textbf{R@1} & \textbf{R@3} & \textbf{R@5} \\
      \midrule
      GPT-4o                            & \textbf{0.3732} & \textbf{0.4243} & \textbf{0.4991} & 0.5221  \\
      Qwen2.5-VL-7B-Instruct            & 0.2349          & 0.3684          & 0.4609          & 0.4859  \\
      Qwen2-VL-7B-Instruct              & 0.1594          & 0.3656          & 0.4590          & 0.4781  \\
      Qwen2-VL-3B-Instruct              & 0.1386          & 0.2663          & 0.3250          & 0.3517  \\
      llava-onevision-qwen2-7b-ov       & 0.1546          & 0.3629          & 0.4497          & 0.4639  \\
      EVA-CLIP (ViT-g)                  & 0.2465          & 0.2654          & 0.4897          & 0.5944  \\
      CLIP (ViT-L/14)                   & 0.2393          & 0.2567          & 0.4885          & \textbf{0.6000} \\
      llava-v1.6-34b                    & 0.1354          & 0.2432          & 0.3779          & 0.4063  \\
      llava-v1.5-7b                     & 0.0991          & 0.2140          & 0.2721          & 0.2973  \\
      llava-v1.6-mistral-7b             & 0.0944          & 0.1932          & 0.2595          & 0.2831  \\
      \bottomrule
    \end{tabular}%
  }
  \label{table:classification_comparison}
\end{table}

\subsection{Task 2. Continual User-Aware Adaptation}  
\label{4-2}
\subsubsection{Task Definition and Baseline Method 1}

Users often exhibit repetitive eating patterns, frequently consuming similar meals over time. 
This suggests that continuously updating a model with newly logged meals could improve classification performance, as previously seen meals become easier to recognize.
Using the “meal date” metadata, we incrementally fine‐tuned a model on all records up to week \(n-1\) and tested on meals logged in week \(n\), as illustrated at the bottom left of Figure~\ref{task}.

Using Qwen2.5‐VL‐3B-Instruct, we evaluated three model variants on held‐out splits: (1) full fine‐tuning, (updates all model parameters), (2) LoRA fine‐tuning (adapts only low-rank adapter weights)~\cite{Hu2022LoRA}, and (3) a zero‐shot baseline with no fine‐tuning.

\subsubsection{Experimental Results 1}

\begin{figure}[tb]
  \centering
  \includegraphics[width=\linewidth]{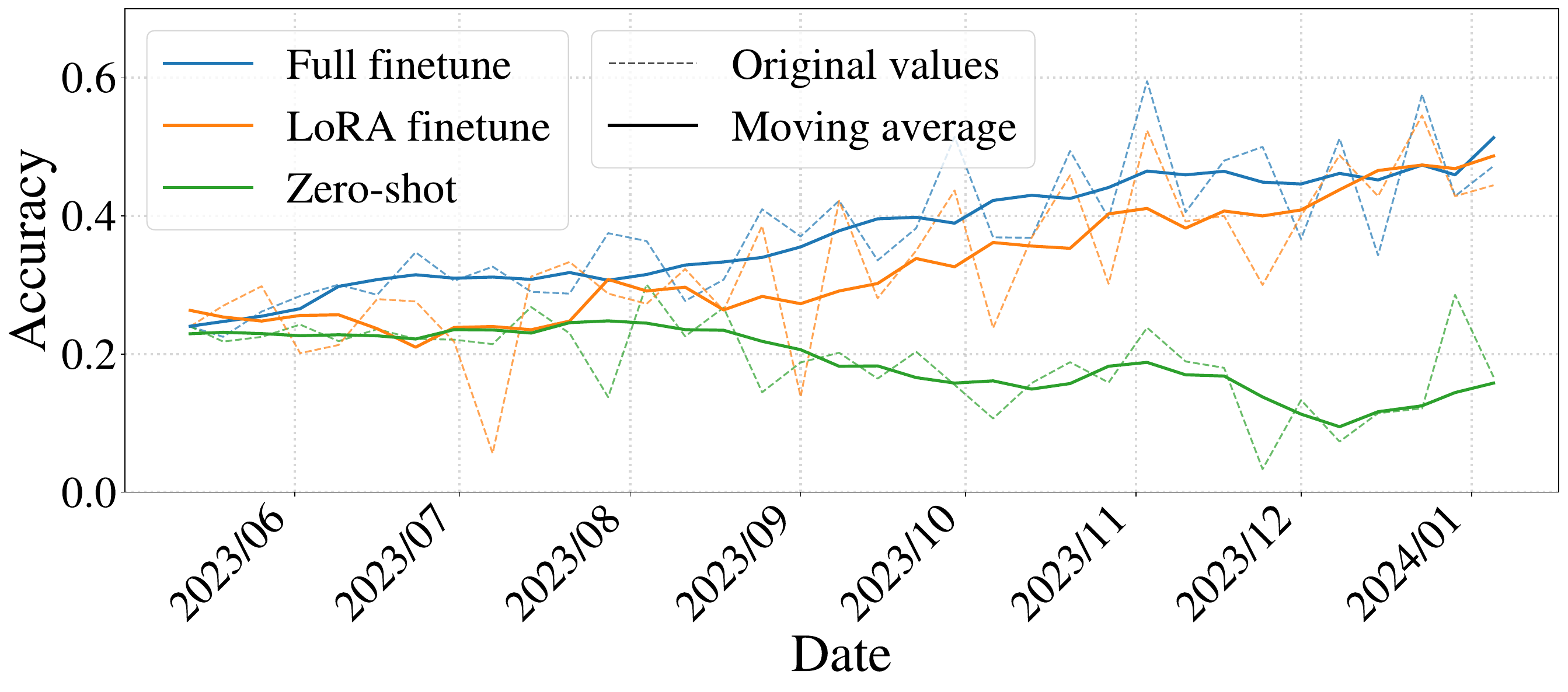}
  \caption{Incremental accuracy: zero-shot, full fine-tuning, and LoRA fine-tuning on all users.}
  \label{fig:incremental}
\end{figure}

Figure~\ref{fig:incremental} shows incremental accuracy for the three variants. Solid lines represent a weighted one-month moving average ($\pm$ two weeks) with weights proportional to the number of weekly records; dotted lines plot the raw weekly accuracy.

Figure~\ref{fig:incremental} shows that with continual fine‐tuning on real meal logs, accuracy climbs from roughly 20\% to over 50\%—a gain of more than 30 percentage points over the zero‐shot baseline—ultimately surpassing even large commercial LMMs. LoRA fine‐tuning, though slightly behind full fine‐tuning in the earliest weeks when only a few dozen records are available, quickly catches up as more data accumulates, achieving near-full-finetune performance with only a small fraction of trainable parameters. In contrast, the zero-shot curve slowly drifts downward, highlighting the importance of users' dietary history updates in counteracting the distributional shift in real-world meal-logging data.

\subsubsection{Task Definition and Baseline Method 2}

To analyze the impact of long-term logging on personalization, we focused on the single most active user (User A), who consistently logged meals from May 12, 2023 to January 12, 2024, submitting 274 images covering 1,003 dish entries. 
On this user’s records, we compared three variants: (1) LoRA fine‐tuned solely on User A’s data, (2) LoRA fine‐tuned on all users, and (3) zero‐shot baseline with no fine‐tuning.

\subsubsection{Experimental Results 2}

\begin{figure}[tb]
  \centering
  \includegraphics[width=\linewidth]{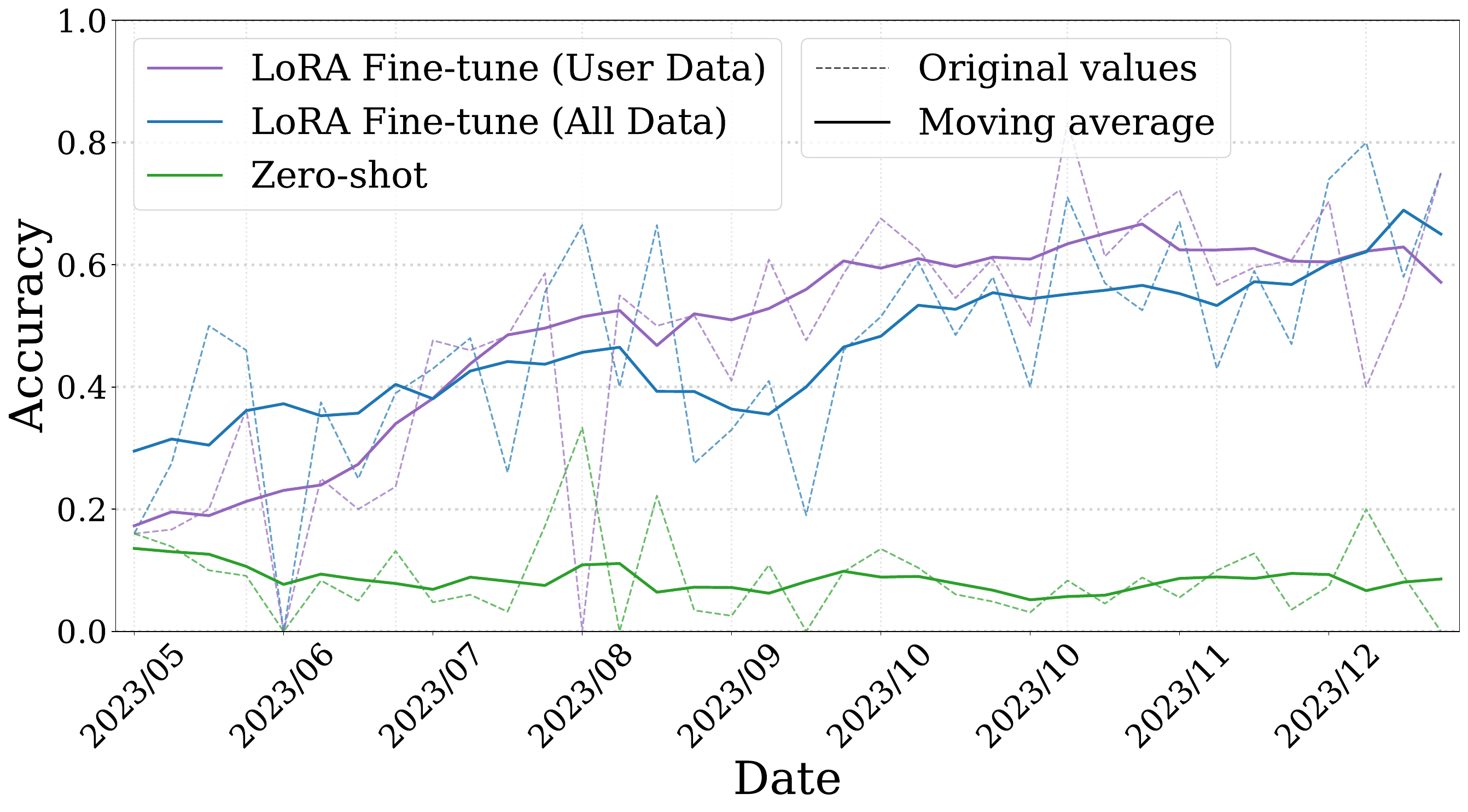}
  \caption{User A accuracy over time: zero‐shot, LoRA fine‐tuned on all users, and LoRA fine‐tuned on User A only.}
  \label{fig:user_personalization}
\end{figure}

Figure~\ref{fig:user_personalization} shows their classification accuracy on User A's records over time. Solid curves represent a one-month moving average ($\pm$ two weeks), while dotted curves plot the raw weekly accuracy.

From Figure~\ref{fig:user_personalization}, both LoRA‐fine‐tuned models ultimately deliver a 50 percentage points increase over the zero‐shot baseline. The all‐data model initially outperforms the user‐specific model, but as more entries from User A accumulate, the personalized model closes the gap—eventually matching or surpassing the all‐data model—demonstrating effective user‐level adaptation.

\subsection{Task 3. Menu-Aware Classification}  
\label{4-3}
\subsubsection{Task Definition and Baseline Methods}

A single photo often contains several dishes. To leverage this context, we first show the whole image to the LMM and ask for a short ``menu'' description summarizing every visible item. Each crop is then re-evaluated with both its own image and the menu description. This two-stage inference uses meal-level cues to distinguish visually similar dishes—e.g., mains vs.\ sides or regional variants—and aims to boost per-dish accuracy. This overall pipeline is illustrated at the bottom right of Figure~\ref{task}.

\subsubsection{Experimental Results}

We used Qwen2.5-VL-7B-Instruct to generate menu descriptions.
Table~\ref{table:desc_guided} shows the model’s F1, Precision@1 (P@1), and Recall@K (R@1, R@3, R@5) when classifying each dish crop with and without a high‐level meal description. 

\begin{table}[tb]
  \centering
  \caption{F1 score, Precision@1 (P@1), and Recall@K (R@1, R@3, R@5) for Qwen2.5‐VL‐7B‐Instruct on the description‐guided classification task.}
    \begin{tabular}{lccccc}
      \toprule
                    & \textbf{F1}  & \textbf{P@1} & \textbf{R@1} & \textbf{R@3} & \textbf{R@5} \\
      \midrule
      w/o description & 0.1594       & 0.1019       & \textbf{0.3656}       & \textbf{0.4590}       & \textbf{0.4781}       \\
      w/ description  & \textbf{0.2304}       & \textbf{0.1788}       & 0.3240       & 0.4032       & 0.4336       \\
      \bottomrule
    \end{tabular}%
  \label{table:desc_guided}
\end{table}

\begin{figure}[tb]
  \centering
  \includegraphics[width=\linewidth]{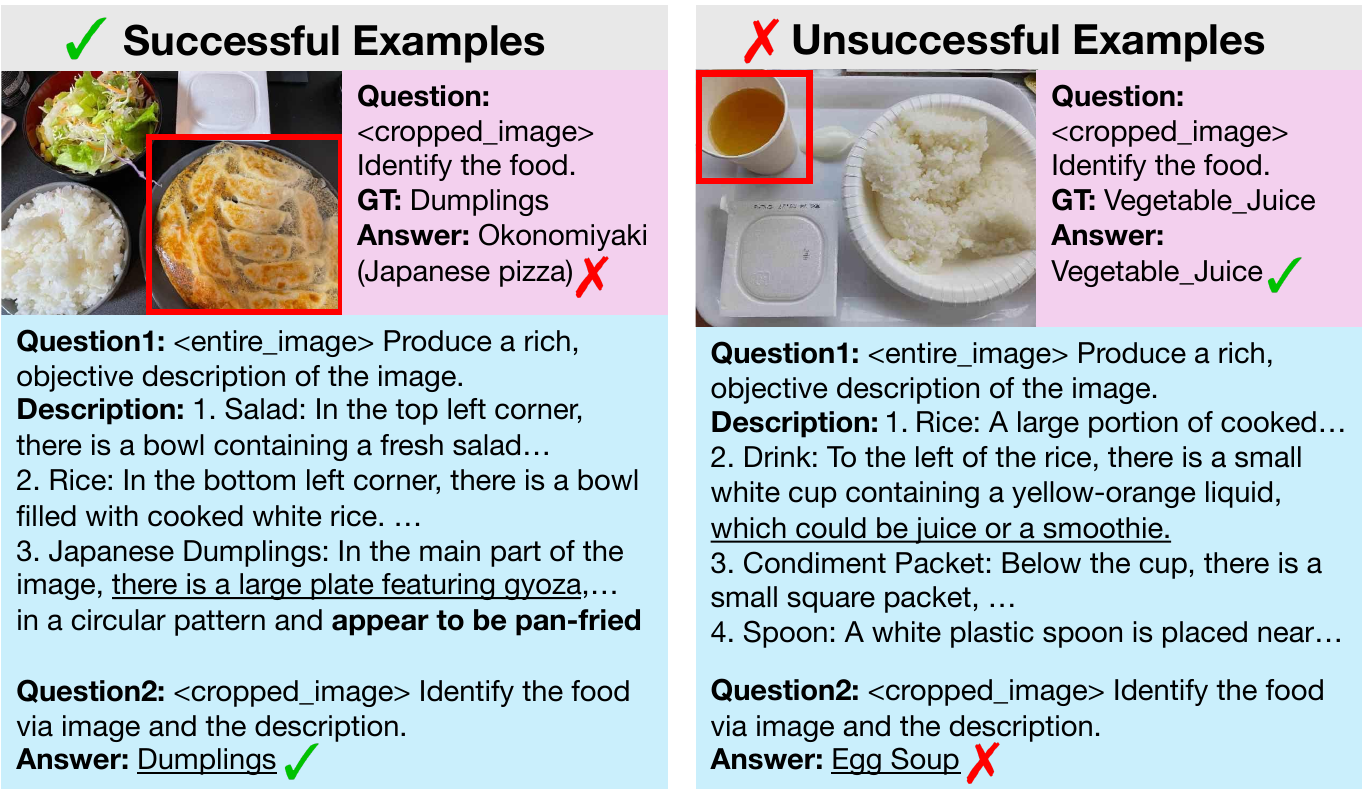}
  \caption{Illustrative examples for the description‐guided classification task. Pink‐shaded regions show the baseline ``without description'' setting, while blue‐shaded regions show the ``with description'' setting.}
  \label{fig:desc_examples}
\end{figure}

Adding the menu description raises F1 from 0.159 to 0.230 and boosts Precision@1 by over 7.5\% (from 0.102 to 0.179), at the cost of a modest drop in Recall. As one illustrative example, the model—rather than defaulting to “Rice” (11.9\% of samples)—now correctly predicts rarer variants like Barley\_Rice, Brown\_Rice, and Fried\_Rice. Crucially, this is not limited to the Rice class: comparable reductions in bias toward other dominant labels and shifts toward their less common subcategories are observed across the dataset, demonstrating a general improvement in label diversity.

Despite this improvement, the addition of menu descriptions also introduces trade-offs: it correctly reclassifies 687 previously misclassified instances but simultaneously misclassifies 1,222 instances that were originally labeled correctly. Figure~\ref{fig:desc_examples} presents representative examples of both successful corrections and new errors.
These results suggest that the LMM effectively utilizes contextual information from the full meal. 
For example, a cropped image of {\it gyoza} is incorrectly classified as {\it okonomiyaki} when evaluated in isolation. However, with the addition of the full-meal description, which mentions rice and side dishes, items typically served with {\it gyoza} but not with {\it okonomiyaki} in Japan, the model correctly predicts the label Dumplings. 
Conversely, some samples that were correctly classified without context become misclassified once the meal description is included, indicating that while context can enhance prediction accuracy, it can also introduce ambiguity or mislead the model in some instances.

\section{Conclusion}

We presented {\it FoodLogAthl-218}, a 6,925‐image, 14,349 bounding box, and 218‐class food dataset derived from real meal logs with rich metadata and natural class imbalance.  Through three tasks—closed‐set classification, continual user‐aware adaptation, and description‐guided classification—we demonstrated that state‐of‐the‐art LMMs struggle with raw food‐log data but improve significantly through incremental and context‐aware fine-tuning.  

\noindent\textbf{Availability.} FoodLogAthl-218 is released for non-commercial academic research. Access requires an institutional e-mail. Request it at \url{https://huggingface.co/datasets/FoodLog/FoodLogAthl-218} via ``Apply''; credentials are issued after a short manual review.

\noindent\textbf{Ethics Statement}
This study was approved by the Ethics Review Committee of the Graduate School of Information Science and Technology, The University of Tokyo (protocol
\textbf{UT-IST-RE-250415-2}).

\section{Acknowledgements}
This work was supported by JSPS KAKENHI Grant Number 23K25247 and JST NEXUS, Japan Grant Number JPMJNX25C9.

\bibliographystyle{ACM-Reference-Format}
\bibliography{sample-base}

\end{document}